\documentclass[10pt,twocolumn,letterpaper]{article}

\usepackage{kotex}
\usepackage[pagenumbers]{cvpr} 

\usepackage{xcolor}
\usepackage{graphicx}
\usepackage{amsmath}
\usepackage{amssymb}
\usepackage{booktabs}
\usepackage{multirow}
\usepackage{verbatim}
\usepackage{enumitem}
\usepackage{makecell}
\usepackage{adjustbox}
\usepackage[accsupp]{axessibility} 

\usepackage[pagebackref,breaklinks,colorlinks]{hyperref}

\usepackage[capitalize]{cleveref}
\crefname{section}{Sec.}{Secs.}
\Crefname{section}{Section}{Sections}
\Crefname{table}{Table}{Tables}
\crefname{table}{Tab.}{Tabs.}

\def\cvprPaperID{7820} 
\def\confName{CVPR}
\def\confYear{2022}

\definecolor{dark_cyan}{RGB}{0,139,139}
\definecolor{gtcolor}{RGB}{0,128,0}

\begin{document}

\title{MatteFormer: Transformer-Based Image Matting via Prior-Tokens}

\author{GyuTae Park$^{1,2}$, SungJoon Son$^{1,2}$, JaeYoung Yoo$^{2}$, SeHo Kim$^{2}$, Nojun Kwak$^{1}$\\
$^{1}$ Seoul National University, South Korea \\
$^{2}$ NAVER WEBTOON AI, South Korea\\
{\tt\small \{gyutae.park, sjson718, yoojy31, seho.kim\}@webtoonscorp.com, nojunk@snu.ac.kr}
}

\maketitle

\begin{abstract}
  In this paper, we propose a transformer-based image matting model called MatteFormer, which takes full advantage of trimap information in the transformer block. Our method first introduces a prior-token which is a global representation of each trimap region (e.g. foreground, background and unknown). These prior-tokens are used as global priors and participate in the self-attention mechanism of each block. Each stage of the encoder is composed of PAST (Prior-Attentive Swin Transformer) block, which is based on the Swin Transformer block, but differs in a couple of aspects:
  1) It has PA-WSA (Prior-Attentive Window Self-Attention) layer, performing self-attention not only with spatial-tokens but also with prior-tokens. 2) It has prior-memory which saves prior-tokens accumulatively from the previous blocks and transfers them to the next block.
  We evaluate our MatteFormer on the commonly used image matting datasets: Composition-1k and Distinctions-646. Experiment results show that our proposed method achieves state-of-the-art performance with a large margin. Our codes are available at \url{https://github.com/webtoon/matteformer}.
\end{abstract}

\section{Introduction}
\label{sec:intro}

{\let\thefootnote\relax\footnotetext{{
Nojun kwak was supported by the NRF (2021R1A2C3006659) and IITP grant (NO.2021-0-01343) funded by the Korea government (MSIT).}}
}

Image matting is one of the most fundamental tasks in computer vision which is mainly used to separate a foreground object precisely for the purpose of image editing and compositing. 
Especially, the foreground not only includes complex objects like human hair and animal fur but also includes transparent objects like glass, bulb and water. Natural image can be represented as a linear combination of foreground $F \in \mathbb{R}^{H\times W \times C}$ and background $B \in \mathbb{R}^{H\times W \times C}$ with alpha matte $\alpha \in \mathbb{R}^{H\times W}$ as follows: 
\begin{equation}
  I_{i} = \alpha_{i}  F_{i} + (1-\alpha_{i}) B_{i}, \quad \alpha_{i} \in [0,1],
  \label{eq:matting}
\end{equation}
where $H, W$ and $C$ denotes the height, the width and the number of channels (3 for a color image) respectively, and $i \in [HW]$ denotes the pixel index. 

In image matting, estimating opacity value $\alpha$ given only observed image $I$, is a highly ill-posed problem if any extra information is not available. Thus, in many works, various types of additional user-inputs (e.g. trimap, scribble, binary mask, background image, etc.) are used, among which, especially trimap is the most common. Trimaps are cost-intensive for users to draw but provide high-quality information about global context such as region information about foreground, background and unknown pixels. Consequently, it is natural to design a model to take a full advantage of this user-input and many works utilizing trimaps have been developed for image matting, most of which are based on convolutional neural networks (CNNs).

\begin{figure}
  \begin{center}
  \includegraphics[width=0.8\linewidth]{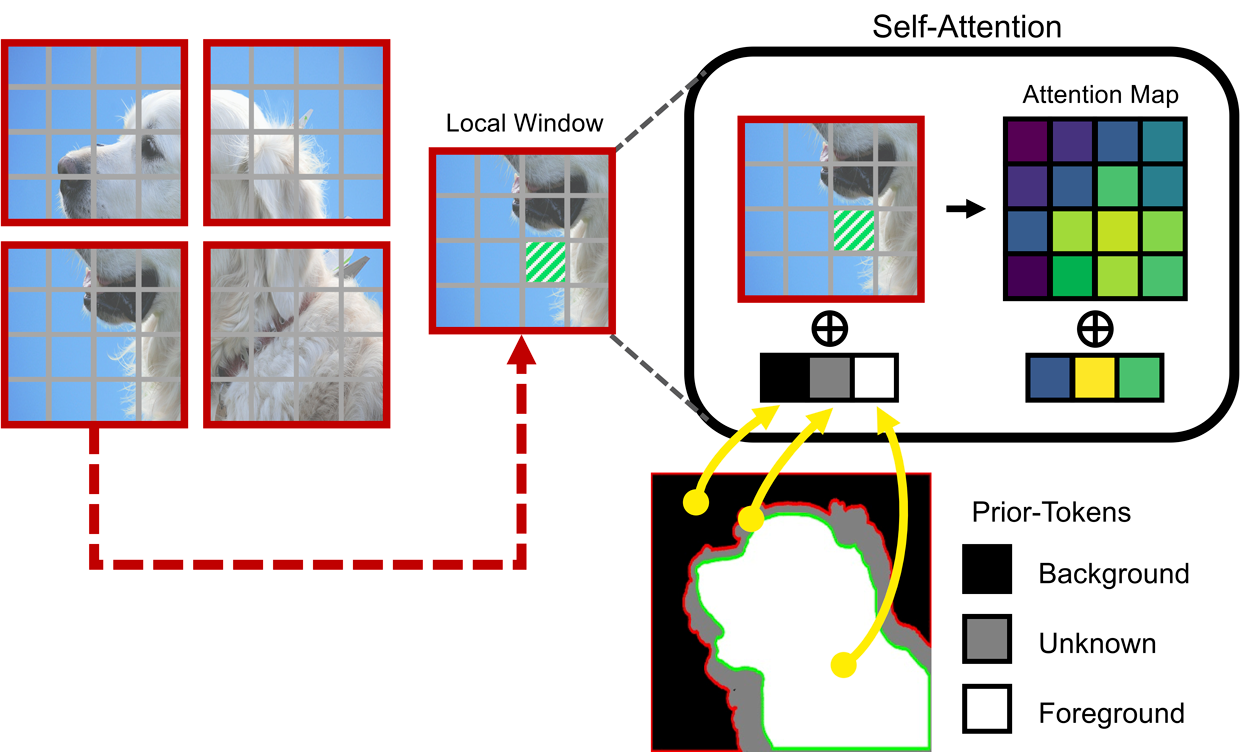}
  \end{center}
  \vspace{-5mm}
  \caption{Prior-tokens are generated via the trimap at the bottom and concatenated to the local spatial-tokens to participate in the self-attention mechanism.}
  \label{fig:intro}
\end{figure} 

While CNNs have been very successful in computer vision tasks, for natural language processing (NLP) tasks, \textit{transformers} earned great success and recently there have been many attempts to introduce transformers in downstream vision tasks as an alternative to CNNs. The seminal work of~\cite{dosovitskiy2020image} proposed Vision Transformer (ViT) and showed impressive performances compare to CNN-based models, demonstrating potential on vision tasks.
However, implementing the global self-attention in ViT needs high computational cost; it is quadratic to the number of patches. To overcome this limitation, several general-purpose transformer backbones~\cite{liu2021swin, dong2021cswin, yang2021focal} have been proposed by reducing computational complexity with local self-attention methods.
For example,~\cite{liu2021swin} proposed a hierarchical transformer structure by introducing the self-attention within local windows to enable linear cost to the input size and the patch merging layer to reduce the number of spatial-tokens in deeper layers. Besides, they proposed the shifting window scheme to exchange information through neighboring windows.
Unfortunately, since the shifting window technique slowly enlarges the receptive field, it is still hard to achieve a sufficiently large receptive field, especially in lower layers.

In this paper, we propose a transformer-based image matting model, named MatteFormer.
We first define a prior-token, which represents global context feature of each trimap region; the foreground, background and unknown region as shown in \cref{fig:intro}.
These prior-tokens are used as global priors and participate in the self-attention mechanism of each block.
The encoder stage is composed of the PAST (Prior-Attentive Swin Transformer) blocks, which are based on the Swin Transformer block~\cite{liu2021swin}.
However, our PAST block is different from the Swin Transformer block in two aspects. First, it has the PA-WSA (Prior-Attentive Window Self-Attention) layer, where self-attention is computed not only with spatial-tokens but also with prior-tokens as shown in \cref{fig:intro}. 
Second, we introduce the prior-memory, memorizing all prior-tokens generated at each block. Through this, prior-tokens from the previous block can be utilized in the PA-WSA layer of the next block.
We evaluate our MatteFormer on Composition-1k and Distinctions-646, which are the commonly used datasets in image matting. The results show that our method achieves state-of-the-art performance.
We also conduct some extensive studies about the effectiveness of prior-tokens, visualization of self-attention maps, usage of ASPP, and the computational cost.

In short, our contributions can be summarized as follows:
\begin{itemize}
\vspace{-2mm}
\item We propose MatteFormer, the first transformer-based architecture for the image matting problem.
\vspace{-2mm}
\item We introduce prior-tokens which imply global information of each trimap region (foreground, background and unknown) and use them as global priors in our proposed network.
\vspace{-2mm}
\item We design the PAST (Prior-Attentive Swin Transformer) block, a variant of the Swin Transformer block, which includes the PA-WSA (Prior-Attentive Window Self-Attention) layer and the prior-memory. 
\vspace{-6mm}
\item We evaluate MatteFormer on Composition-1k and Distinctions-646, showing that our method achieves state-of-the-art performance with a large margin.
\end{itemize}

\begin{figure*}
  \begin{center}
  \includegraphics[width=0.75\linewidth]{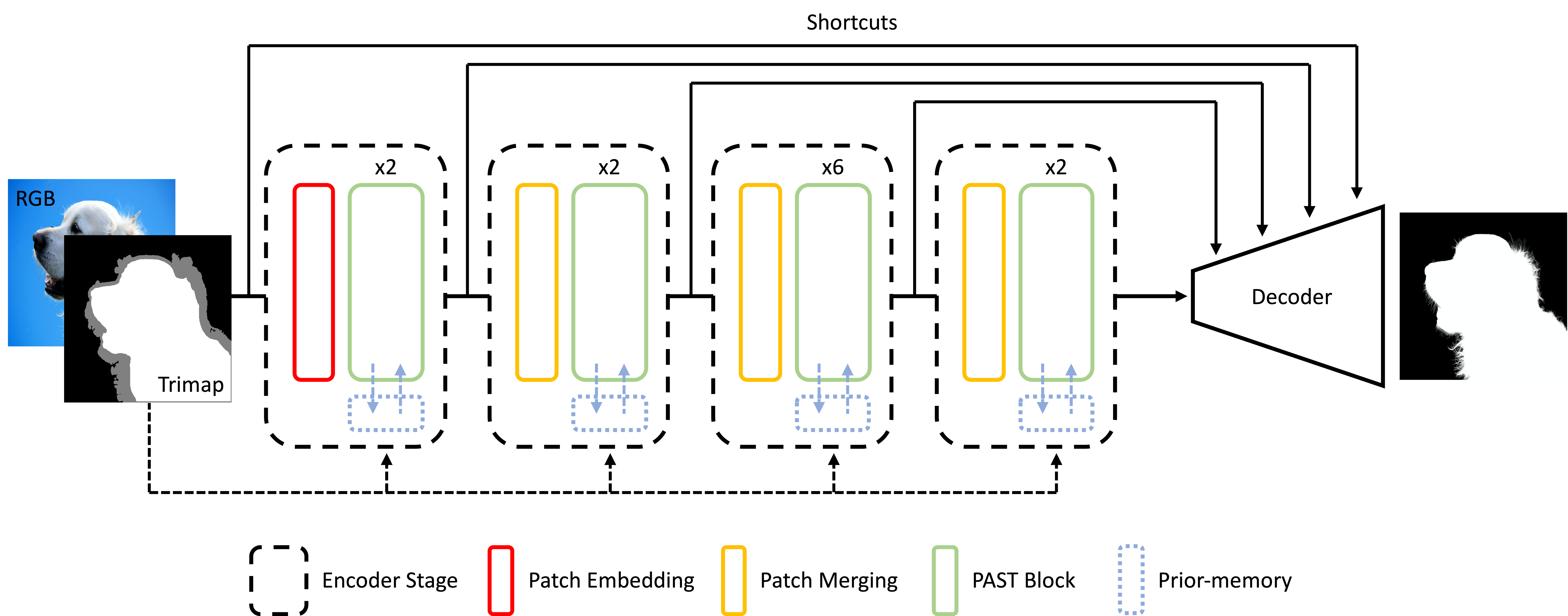}
  \end{center}
  \vspace{-5mm}
  \caption{Overall architecture of our proposed MatteFormer, which has a simple encoder-decoder structure with shortcut connections. Each encoder stage includes the proposed PAST (Prior-Attentive Swin Transformer) block. The trimap contributes to generating prior-tokens in each PAST block. The prior-tokens are stored at the prior-memory for usage at later blocks.}
  \label{fig:MatteFormer}
\end{figure*}

\section{Related Works }
\label{sec:related works}

\subsection{Natural Image Matting}

\textbf{Traditional methods}.
As natural image matting is considered as an ill-posed problem, most of matting algorithms make use of user-input as an additional prior. The \textit{trimap} is frequently used which provides information about foreground, background and unknown regions.
Traditional methods predict the alpha matte by primarily utilizing color features. They are mainly divided into sampling-based and propagation-based methods, according to the way of using the color features.

Sampling-based methods~\cite{gastal2010shared, chuang2001bayesian, he2011global, shahrian2013improving, wang2007optimized} make use of color similarities between unknown regions and known (foreground and background) regions to estimate the alpha matte with foreground and background color statistics.
On the other hand, propagation-based methods~\cite{chen2013knn, lee2011nonlocal, levin2007closed, levin2008spectral, he2010fast, sun2004poisson}, also known as affinity-based methods, propagate the alpha values from a known region (foreground and background) to the unknown region to estimate alpha matte via the affinities of neighboring pixels.

\textbf{Learning-based methods}. 
Since traditional approaches highly depend on color features, these may produce artifacts due to lack of semantic information. Like deep-learning has achieved great success on many other vision tasks, image matting performance also has been significantly improved through CNNs. Meanwhile, because drawing a trimap requires expensive labor cost, approaches without trimap also have been studied.

For the trimap-based methods,
\cite{xu2017deep} proposed a two-stage architecture and released the Composition-1K dataset.
\cite{lutz2018alphagan} utilized a GAN (generative adversarial network) framework to improve performance.
\cite{tang2019learning} mixed the sampling-based method with deep learning method.
\cite{hou2019context} designed two encoders for local feature and global context, estimating both foreground and alpha matte.
\cite{lu2019indices} proposed an index-guided encoder-decoder structure with the concept of learning to index.
\cite{li2020natural} developed a guided contextual attention module propagating high-level opacity globally based on low-level affinity.
\cite{yu2020high} proposed a patch-based method for high-resolution inputs with a module addressing cross-patch dependency and consistency issues between patches.
\cite{liu2021towards} designed the model with the textural compensate path to enhance fine-grained details.
\cite{sun2021semantic} used the class of matting patterns and proposed the semantic trimap.

For the trimap-free methods,
~\cite{zhang2019late, qiao2020attention} predicted alpha matte with a single RGB only.
~\cite{zhang2019late} proposed structure of two decoders for classifying foreground and background, and fused them in the extra network.
~\cite{qiao2020attention} designed hierarchical attention structure and  proposed Distinctions-646 dataset.
~\cite{sengupta2020background, lin2021real} proposed the method of using an additional background image instead of the trimap.
~\cite{yu2021mask} used binary mask as additional input and proposed a method to progressively refine the uncertain regions through the decoding process.

Since trimap plays a role as a strong hint, the trimap-based methods generally perform better than the trimap-free methods. However, many algorithms have used the trimap by simply concatenating it with input RGB channels, which do not fully utilize the potential of trimap. 
In this respect, we follow the trimap-based method and propose method of fully making use of the trimap in networks via prior-tokens.

\subsection{Transformer on Vision Tasks}

Solely based on self-attention mechanism, \textit{transformer}~\cite{vaswani2017attention} has shown great success on NLP (natural language processing) tasks and has been a base structure for most language models. Besides,~\cite{dosovitskiy2020image} applied the transformer to the image classification task demonstrating that transformer-based architectures can achieve competitive performance as an alternative to CNN. 

With the potential of ViT~\cite{dosovitskiy2020image} and its follow-ups~\cite{touvron2021training, yuan2021tokens, chu2021we, han2021transformer}, researchers have studied applying transformers to typical vision tasks.
For example, numerous transformer-based models have been proposed on image classification~\cite{liu2021swin, wu2020visual, li2021localvit, liu2021transformer, vaswani2021scaling, zhang2021multi, dong2021cswin, wu2021cvt, ryoo2021tokenlearner}, object detection~\cite{carion2020end, beal2020toward, liu2021swin, wang2021anchor}, semantic segmentation~\cite{zheng2021rethinking, xie2021segformer, liu2021swin}, image completion~\cite{wan2021high} and low-level vision tasks~\cite{chen2021pre, liang2021swinir, lu2021efficient}.

Meanwhile, many researches~\cite{wang2021pyramid, liu2021swin, zhang2021multi, dong2021cswin, yang2021focal, wu2021cvt} have been conducted on designing general-purpose transformer backbone for downstream vision tasks.
Especially,~\cite{liu2021swin, dong2021cswin, yang2021focal} focus on variation of self-attention for reducing computational cost and on hierarchical structures for extracting multi-scale features.
For example, Swin Transformer~\cite{liu2021swin}, which is used as our baseline, performs self-attention in a local window with the shifted window scheme allowing for cross-window connections. However, it has a limitation of still-insufficient receptive field, especially in lower layers. Our approach addresses this issue via prior-tokens and achieves state-of-the-art performance on image matting problem.

\section{Methodology}
\subsection{Network Structure}
\label{sec:structure}
As shown in \cref{fig:MatteFormer}, our proposed network called MatteFormer has a typical encoder-decoder structure with shortcut connections. Each encoder stage is composed of a PAST (Prior-Attentive Swin Transformer) block which is a variant version of the Swin Transformer block, and the patch merging layer which reduces the number of tokens. Our proposed PAST block compensates for the insufficient receptive field, originating from the self-attention within local windows, via prior-tokens. In the decoder, we use a quite simple structure as used in\cite{yu2021mask} with a few convolution and upsample layers. Intermediate features in the encoding layers are delivered to the corresponding decoder layer in a direct way by shortcut connections.

\subsection{Prior-token}
\label{sec:prior-token}
As a trimap includes definite region information, it is natural to design the network to make full use of this powerful hint, for achieving better results. We first generate prior-tokens of three query regions; foreground, background and unknown areas, by averaging all tokens belonging to the corresponding query region. For example, in \cref{fig:intro}, the foreground prior-token is the mean feature of all spatial-tokens lying on the foreground area (white trimap region in the figure).
Specifically, the prior-token $\mathbf{p}^q$ can be formulated as
\begin{equation}
  \mathbf{p}^q = \frac{1}{N_{q}}\sum ^{N}_{i=1} r^{q}_{i} \cdot \mathbf{z}_i, \quad q \in \{ \text{fg, bg, uk} \}, 
  \label{eq:prior}
\end{equation}

where $q$ is the query which takes a value among foreground (fg), background (bg) or unknown (uk), $i$ is the token index and $\mathbf{z}_i$ denotes the feature of a spatial-token. $r_{i}^{q}$ is a binary value depending on whether the $i$-th token is on the corresponding query region ($r_i^q = 1$) or not ($r_i^q = 0$). 

For example, $r_{i}^{fg}$ is 1 if the $i$-th spatial-token is in the foreground region. $N_{q} = \sum_{i}^N r_i^q$ is the number of tokens belonging to the region $q$, indicating how many spatial-tokens are in the corresponding query region. $N$ denotes the total number of spatial-tokens.
For the sake of easy computation, if the area corresponding to a spatial-token has more than one query region, i.e, the area is in the boundary, the spatial-token is allocated to the dominant query region.
Finally, a prior-token $\mathbf{p}^q$ is assumed to have informative representation as a global prior for the corresponding query region.

\begin{figure}[t]
  \begin{center}
  \includegraphics[width=1.0\linewidth]{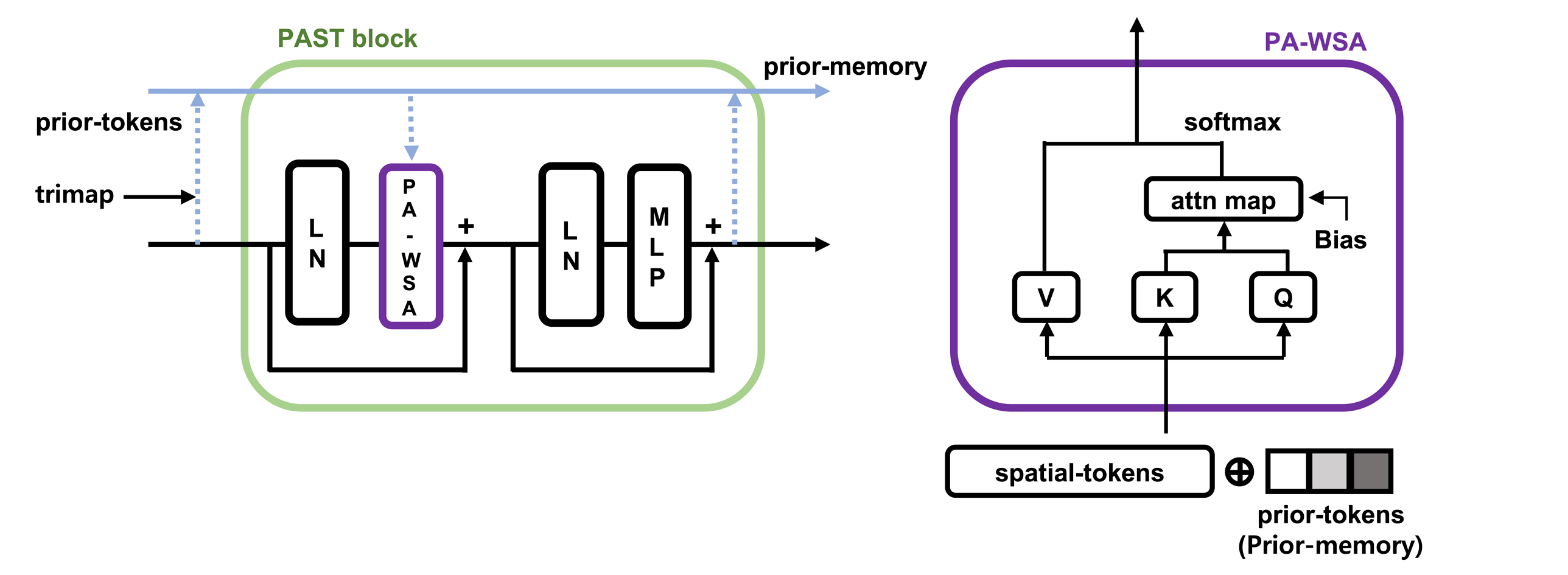}      
  \end{center}
  \vspace{-5mm}
  \caption{Proposed PAST (Prior-Attentive Swin Transformer) block (left). It is based on the Swin Transformer block but differs in two aspects. One is that not only spatial-tokens in the local window but also prior-tokens are used in the PA-WSA (Prior-Attentive Window Self-Attention) layer (right). The other is that it has a prior-memory.}
  \label{fig:PAST}
\end{figure}

\subsection{Prior-Attentive Swin Transformer Block}
\label{sec:PAST}

In this work, we use Swin Transformer~\cite{liu2021swin} as our base model, which showed great promise on mainstream vision tasks such as image classification, object detection and semantic segmentation with a hierarchical architecture design and the shifted window scheme. It has a great efficiency with two aspects. One is that local self-attention on non-overlapping windows enables for model to cover large image size with efficient computation. The other is that shifted window partitioning allows for cross-window connection with enhancing long-range dependency.
However, it has a limitation that shifted windows enlarge the receptive field slowly, which leads to an insufficient attention region, especially in lower layers.

To address the issue, we propose a Prior-Attentive Swin Transformer (PAST) block as shown in \cref{fig:PAST}. Our PAST block is based on the Swin Transformer block but different from it in two aspects; 
First, in the self-attention layer, a token can attend not only to spatial-tokens in the local window but also to global features represented by prior-tokens. Specifically, three prior-tokens are used in our method; foreground, background and unknown prior-tokens. Second, it has a prior-memory, which stores prior-tokens generated from the previous block. Accumulated prior-tokens are used in the next block as informative priors when calculating the self-attention.

\textbf{Self-attention with prior-tokens.}
The detailed structure of our PAST block is as follows: First, prior-tokens generated in \cref{sec:prior-token} are concatenated with spatial-tokens in a local window. With these global prior-tokens and local spatial-tokens, self-attention is performed in the Prior-Attentive Window Self-Attention (PA-WSA) layer with matrix multiplications of queries, keys and values in a multi-head manner.
Specifically, the self-attention mechanism is as follows
\begin{equation}
  Attention(Q,K,V) = SoftMax(Q K^{T} \times s + B)V
  \label{eq:self-attention}
\end{equation}
where $ K, V \in \mathbb{R}^{(M^{2}+{N_{p}}) \times d} $, $ Q \in \mathbb{R}^{M^{2} \times d} $ are $key$, $value$ and $query$ matrices respectively.
$d$ is the feature dimension of $key$, $value$ and $query$. 
$M$ is the window size, thus $M^{2}$ is the number of tokens in a local window. ${N_{p}}$ is the number of prior-tokens and $s$ is a scaling factor. 
$B$ is a relative position bias slightly modified from~\cite{liu2021swin}. 
Between spatial-tokens in a local window, relative positions lies on range $\left [-M+1, M-1\right ]$ both along x- and y-axis.
In our case, as we use extra prior-tokens, we set an auxiliary matrix $\hat{B} \in \mathbb{R}^{{\left ( 2M-1 \right )^{2} + {N_{p}}}}$. Values of $B \in \mathbb{R}^{M^{2} \times (M^{2}+{N_{p}})}$ are taken from $\hat{B}$ and the relative position bias $B$ adjusts the values of attention map according to the relative position.
In this way, a query spatial-token can attend to both local spatial-tokens and global prior-tokens.

\textbf{Prior-memory.}
Prior-tokens generated from different PAST blocks have different representations for the corresponding query region. That is, the prior-tokens from the previous blocks can deliver informative context to the current block. To implement this, we first define a prior-memory. Like local spatial-tokens, prior-tokens sequentially pass through the PA-WSA, normalization and MLP layer. After that, the prior-tokens are appended to the prior-memory. In the next block, accumulated tokens are used as priors in the PA-WSA layer, i.e, the later block has larger $N_p$ than the former blocks. 
More specifically, the $b$-th block in each of the four stages in \cref{fig:MatteFormer} has $3\times b$ prior-tokens in the corresponding prior-memory.

\subsection{Training Scheme}
\label{sec:Training}

The total loss function is defined as a weighted sum of three loss functions; $L_{l1}$ loss, composition loss\cite{xu2017deep} and Laplacian loss\cite{hou2019context}, like in\cite{yu2021mask}.
\begin{equation}
  L_{total} = L_{l1} + L_{comp} + L_{lap}
  \label{eq:loss}
\end{equation}
where $L_{l1}$ is the absolute difference between the ground truth alpha and the predicted alpha. $L_{comp}$ indicates the absolute difference between the ground truth image and the composited image, which is calculated from \cref{eq:matting} with the ground truth foreground, background and predicted alpha mattes. $L_{lap}$ measures the differences of Laplacian pyramid representations of the alpha maps and captures the local and global difference. 

In the decoding process, we use PRM (Progressive Refinement Module)~\cite{yu2021mask} to generate a precise output map as a coarse-to-fine manner.
First, the decoder outputs three alpha mattes from different intermediate layers whose output size is 1/8, 1/4 and 1/1 of an input resolution respectively, and then resized to input resolution.
Next, through the PRM, outputs are selectively fused and uncertain regions are progressively refined. Specifically, for a current output index $l$, a refined alpha map $\alpha_{l} \in \mathbb{R}^{H\times W}$ is calculated with raw matting output ${\alpha_{l}}'\in \mathbb{R}^{H\times W}$ and self-guidance mask $g_{l}\in \mathbb{R}^{H\times W}$ as follows:
\begin{equation}
  \alpha_{l} = {\alpha_{l}}' \odot g_{l} + \alpha_{l-1} \odot (1-g_{l}),
  \label{eq:PRM1}
\end{equation}
\begin{equation}
    g_{l}(x,y) = 
\begin{cases}
    1, & \text{if } 0 <\alpha_{l-1}(x,y) < 1 \\
    0,              & \text{otherwise} \\
\end{cases},
\label{eq:PRM2}
\end{equation}
where $\odot$ denotes element-wise multiplication.
The self-guidance mask $g_{l}$ is obtained from the previous alpha map $\alpha_{l-1}$.
It is defined to have 0 if predicted pixel is definite region (foreground or background), and 1 if transparent region.
The non-confident pixels in the previous output $\alpha_{l-1}$ are replaced with pixels of the current output ${\alpha_{l}}'$ according to $g_{l}$. Meanwhile, the confident pixels of $\alpha_{l-1}$ are not updated.
In this way, confident regions are preserved and the current output can focus only on refining the non-confident region. 

Our data augmentation setting is set similar to that of~\cite{li2020natural} and~\cite{yu2021mask}. We first perform an affine transformation with a random degree, scale and flip. Then we randomly crop both the image and the trimap to a fixed size. After that, random color jittering is applied. Finally, the augmented foreground is composited with a background image.

\section{Experiments}

In this section, We evaluate our MatteFormer on two public datasets, Composition-1k~\cite{xu2017deep} and Distinctions-646~\cite{qiao2020attention}, which are commonly used in the image matting task. We first describe the experimental environments; datasets, evaluation metrics and implementation details. Next, we compare the results of our MatteFormer to those of other state-of-the-art works. Finally, we conduct some ablation studies on our proposed method.

\subsection{Datasets and Evaluation}
Composition-1k provides unique 50 foreground images with the corresponding ground truth alpha mattes as the test set. Background test images are pre-defined from PASCAL VOC2012~\cite{everingham2010pascal}. By compositing the foreground and background images, the number of test samples is 1,000 in total. The train set consists of 431 foreground object images with ground truth alpha mattes. Contrast to the test set, train background images are sampled from MS COCO~\cite{lin2014microsoft}.

Distinctions-646 is comprised of 646 distinct foreground images and it has more versatility and robustness than Composition-1k. Foreground samples are divided into 596 train and 50 test samples. Like Composition-1k, test background samples are pre-defined with PASCAL VOC2012. Foreground images are synthesized with background images following the same composition rule in~\cite{xu2017deep}.
Unfortunately, as distinctions-646 does not release official trimaps unlike Composition-1k, a fair comparison with the previous works is hard.

We evaluate our MatteFormer using four major quantitative metrics in image matting: sum of absolute differences (SAD), mean square error (MSE), slope (Grad), and connectivity (Conn). We use official evaluation code provided by~\cite{xu2017deep}. Note that a model with lower metric values can predict more precise alpha mattes.

\subsection{Implementation Details}

We implement PAST block based on the Swin Transformer block. 
We introduce prior-tokens to participate in the self-attention mechanism with local spatial-tokens and feed them to other layers in the same way as spatial-tokens. A prior-memory is put on each stage, memorizing prior-tokens from only the PAST blocks in the same stage for simplicity.

Our encoder is first initialized with the Tiny model of Swin Transformer pretrained on ImageNet~\cite{deng2009imagenet}, then trained on the image matting dataset in an end-to-end manner.
Since we use a trimap and an RGB image as network input, the number of input channels is 6 which is different from that of the pretrained model.
Thus, we bring the weight of the pretrained patch-embedding layer only to the first 3 channels (RGB) of our patch-embedding layer.
As the size of our relative position bias table $\hat{B}$, is bigger than that of the pretrained model due to prior-tokens, we bring the pretrained bias table to the front of our bias table weight. 
The shortcut layers which deliver the encoder features to the decoder layers are composed of 3$\times$3 convolutions with normalization layers. As a decoder, we simply followed~\cite{yu2021mask} which use a simple CNN-based structure with 3$\times$3 convolution layers and upsample layers. Both shortcut and decoder layers are randomly initialized.

When training, we set the network input size to 512x512 and the batch size to 20 in total on 2 GPUs.
The learning rate is initialized to $4 \cdot 10^{-4}$. We use Adam optimizer with $\beta_{1}=0.5$ and $\beta_{2}= 0.999$.

\subsection{Results on Image Matting Datasets}

\begin{figure*}
  \captionsetup[subfigure]{labelformat=empty}
  \centering
  \begin{subfigure}{0.14\textwidth}
    \includegraphics[width=\linewidth]{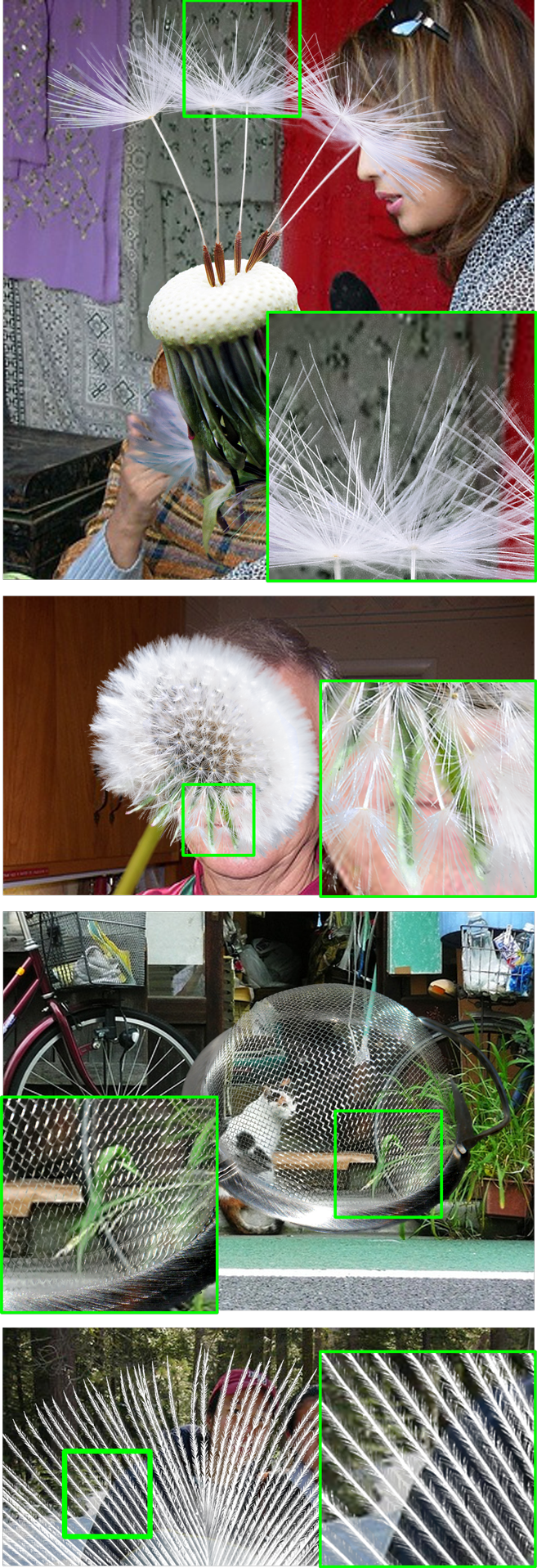}      
    \caption{Image}
    \label{fig:qual_image}
  \end{subfigure}%
  \begin{subfigure}{0.14\textwidth}
    \includegraphics[width=\linewidth]{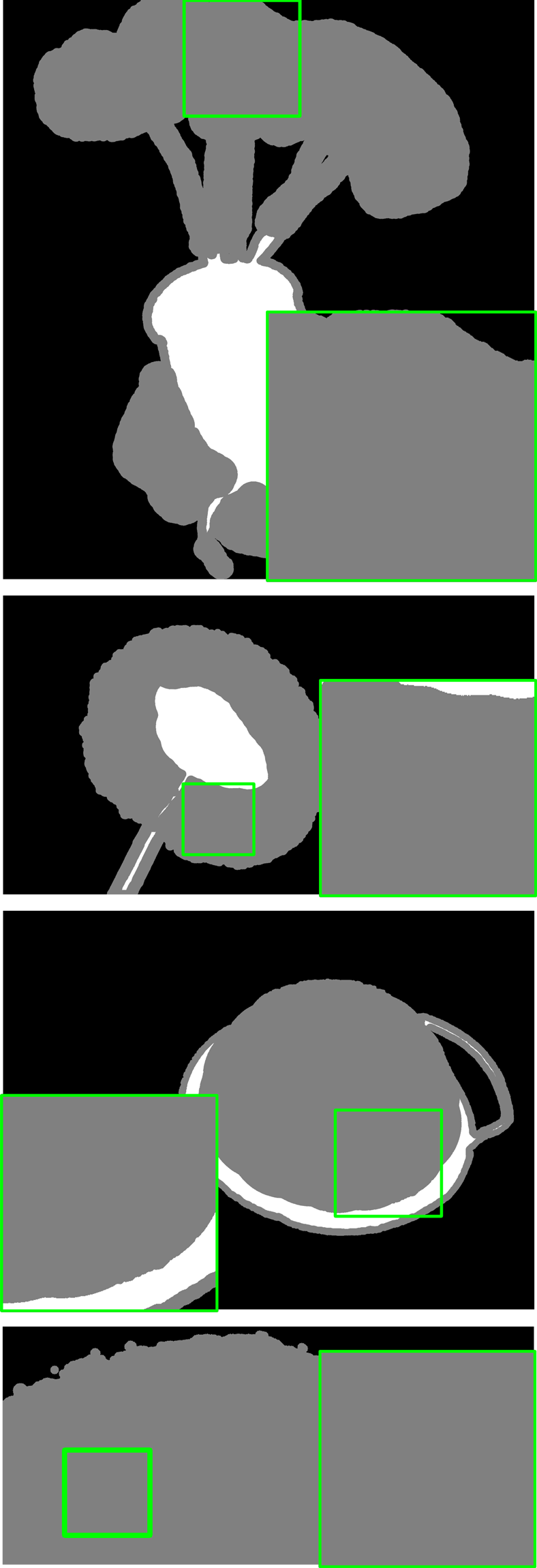}      
    \caption{Trimap}
    \label{fig:qual_trimap}
  \end{subfigure}%
  \begin{subfigure}{0.14\textwidth}
    \includegraphics[width=\linewidth]{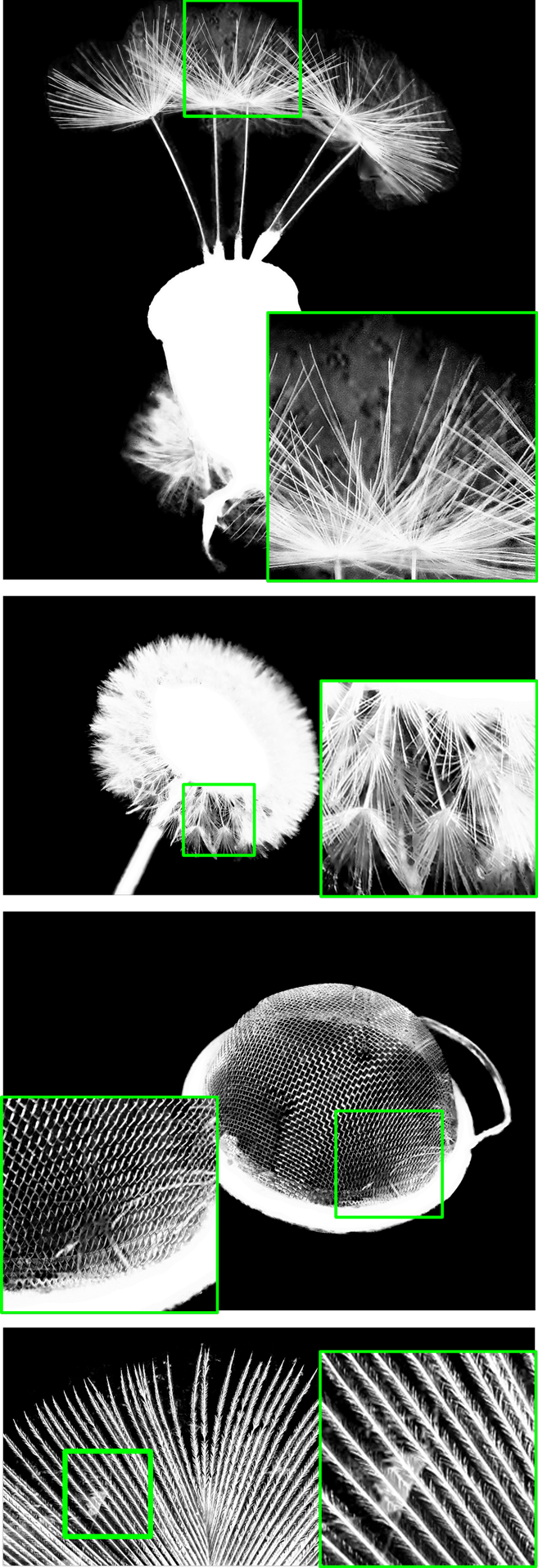}      
    \caption{IndexNet\cite{lu2019indices}}
    \label{fig:qual_index}
  \end{subfigure}%
  \begin{subfigure}{0.14\textwidth}
    \includegraphics[width=\linewidth]{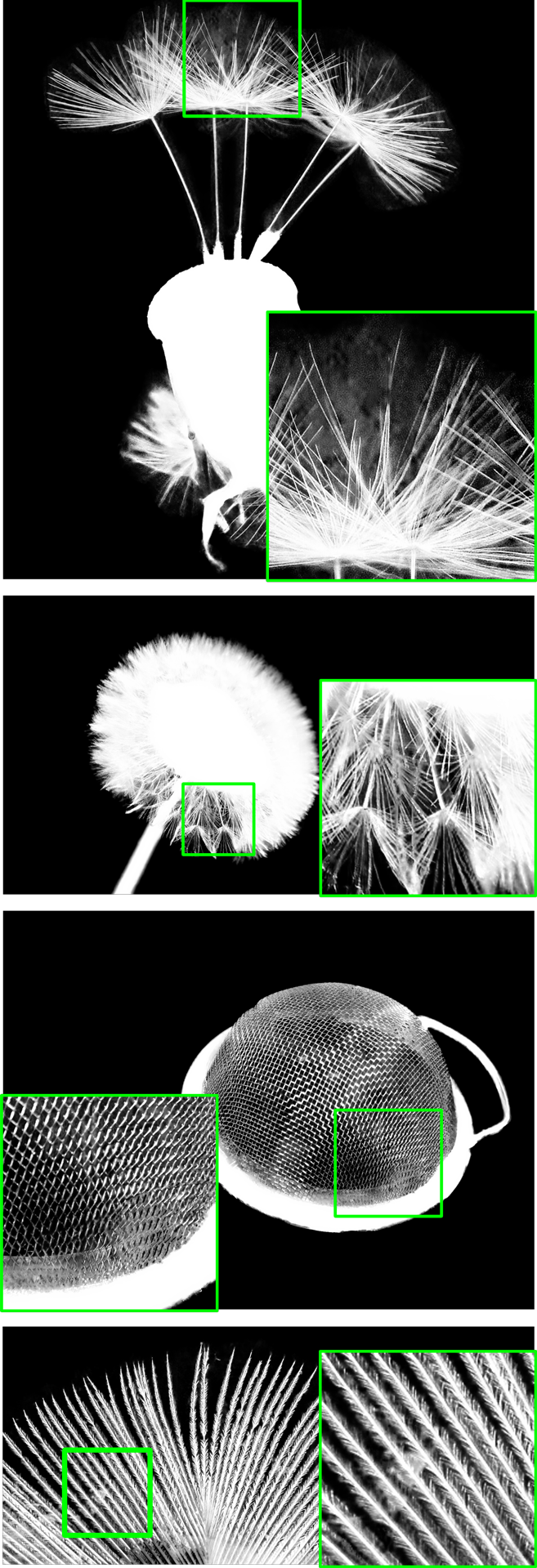}      
    \caption{GCA Matting\cite{li2020natural}}
    \label{fig:qual_gca}
  \end{subfigure}%
  \begin{subfigure}{0.14\textwidth}
    \includegraphics[width=\linewidth]{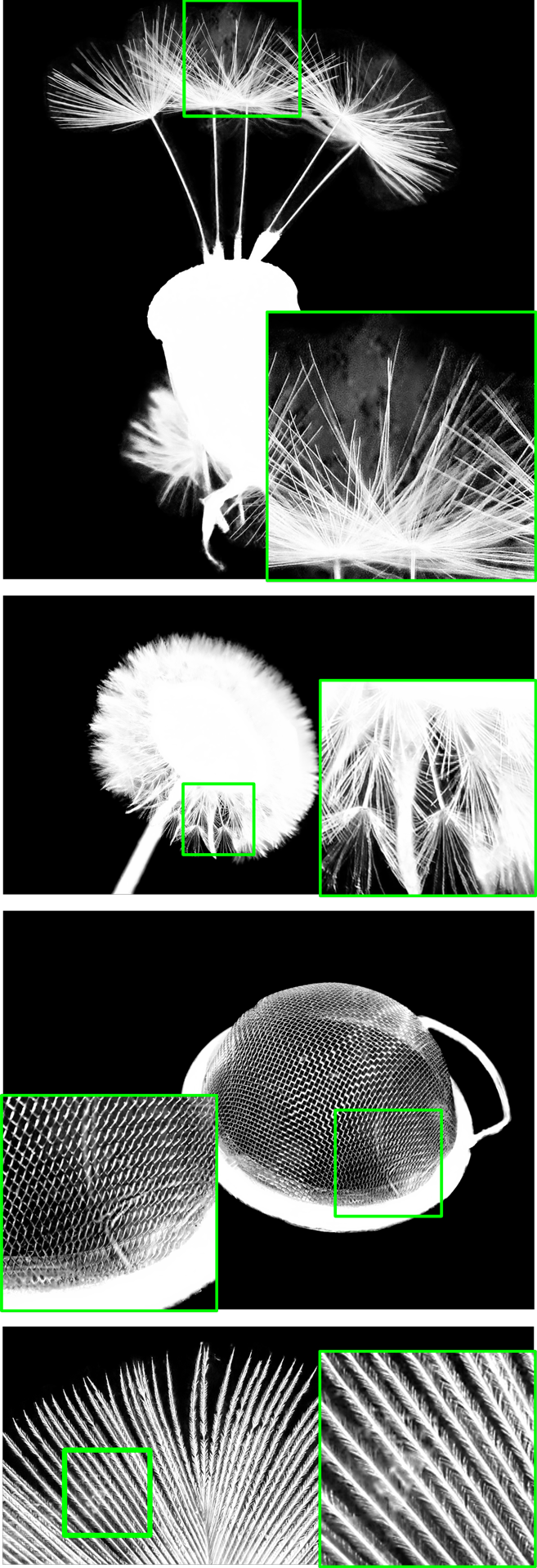}      
    \caption{MG Matting\cite{yu2021mask}}
    \label{fig:qual_mg}
  \end{subfigure}%
  \begin{subfigure}{0.14\textwidth}
    \includegraphics[width=\linewidth]{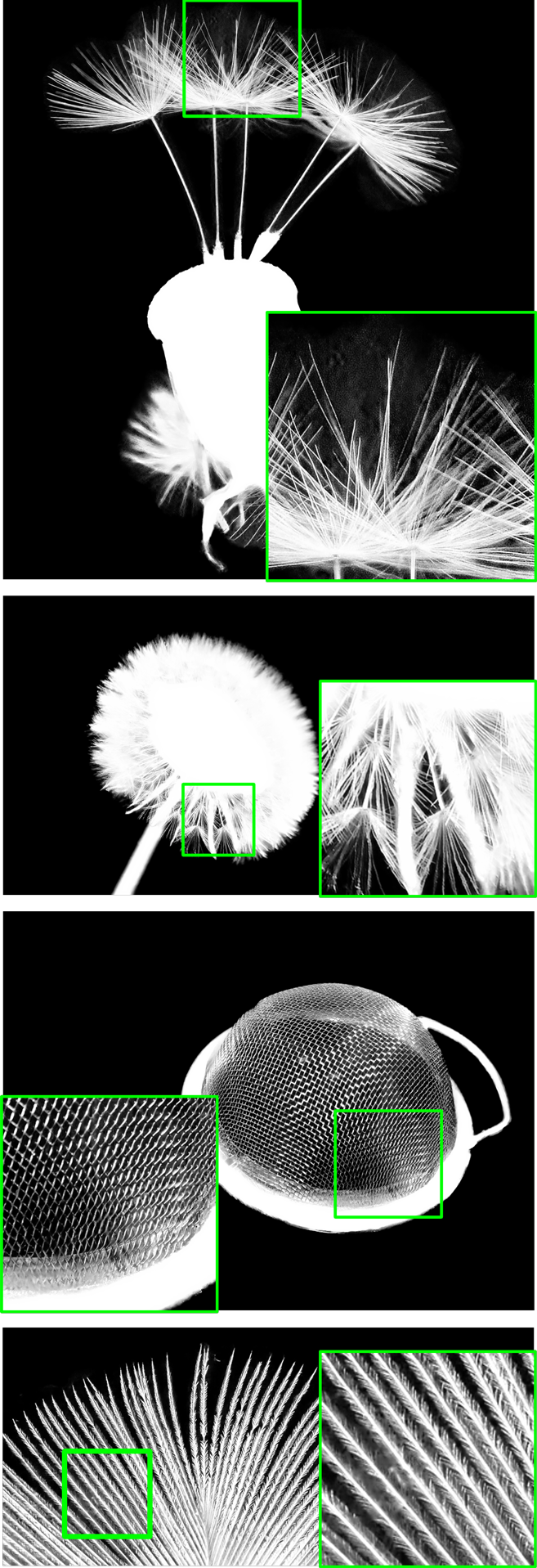}      
    \caption{MatteFormer (Ours)}
    \label{fig:qual_mf}
  \end{subfigure}%
  \begin{subfigure}{0.14\textwidth}
    \includegraphics[width=\linewidth]{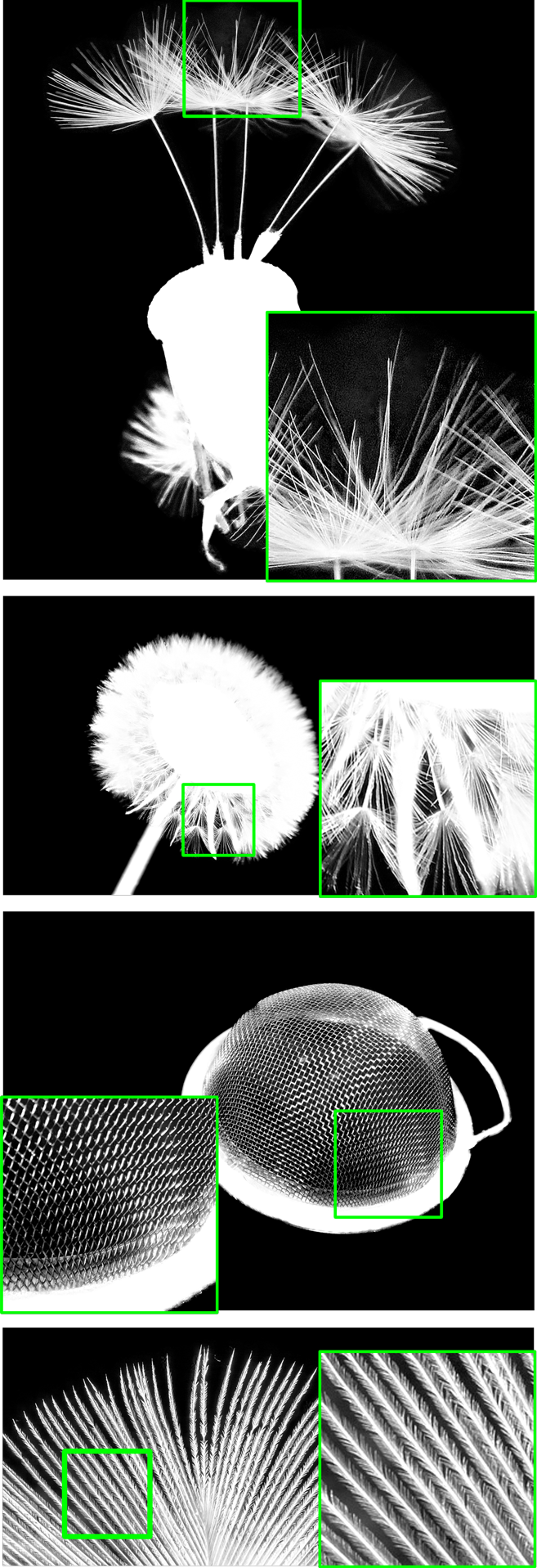}      
    \caption{GT}
    \label{fig:qual_gt}
  \end{subfigure}%
  \vspace{-2mm}
  \caption{The qualitative comparison results on Composition-1k. Best viewed by zooming in.}
  \label{fig:qualitative_results}
\end{figure*}

\begin{table}[t]
  \centering
  \begin{adjustbox}{width = \linewidth}
  \begin{tabular}{l | c c c c}
    \toprule
    Method & SAD & \makecell{MSE \\ ($10^{-3}$)} & Grad & Conn\\
    \midrule
    Learning Based Matting \cite{zheng2009learning} & 113.9 & 48 & 91.6 & 122.2 \\
    Closed-Form Matting \cite{levin2007closed} & 168.1 & 91 & 126.9 & 167.9 \\
    KNN Matting \cite{chen2013knn} & 175.4 & 103 & 124.1 & 176.4 \\
    Deep Image Matting \cite{xu2017deep} & 50.4 & 14 & 31.0 & 50.8 \\
    AlphaGan \cite{lutz2018alphagan} & 52.4 & 30 & 38.0 & - \\
    IndexNet \cite{lu2019indices} & 45.8 & 13 & 25.9 & 43.7 \\
    HAttMatting \cite{qiao2020attention} & 44.0 & 7.0 & 29.3 & 46.4 \\
    AdaMatting \cite{cai2019disentangled} & 41.7 & 10.0 & 16.8 & - \\
    SampleNet \cite{tang2019learning} & 40.4 & 9.9 & - & - \\
    Fine-Grained Matting \cite{liu2021towards} & 37.6 & 9.0 & 18.3 & 35.4 \\
    Context-Aware Matting \cite{hou2019context} & 35.8 & 8.2 & 17.3 & 33.2 \\
    GCA Matting \cite{li2020natural} & 35.3 & 9.1 & 16.9 & 32.5 \\
    HDMatt \cite{yu2020high} & 33.5 & 7.3 & 14.5 & 29.9 \\
    MG Matting \cite{yu2021mask} & 31.5 & 6.8 & 13.5 & 27.3 \\
    MG Matting-trimap* & 28.9 & 5.7 & 11.4 & 24.9 \\
    MG Matting-trimap,res50* & 28.4 & 5.4 & 11.1 & 24.3 \\
    TIMINet \cite{liu2021tripartite} & 29.1 & 6.0 & 11.5 & 25.4 \\
    SIM \cite{sun2021semantic} & 28.0 & 5.8 & 10.8 & 24.8 \\
    \hline
    \bf{Ours (MatteFormer)} & \bf{23.8} & \bf{4.0} & \bf{8.7} & \bf{18.9} \\
    \bottomrule
  \end{tabular}
  \end{adjustbox}
  \caption{Results on Composition-1k test set. * denotes the baseline of comparison which is our reproduction of MG Matting with trimap input. Original MG Matting is based on ResNet-34 and MG Matting-trimap,res50* uses ResNet-50.}
  \label{tab:result_composition1k}
  \vspace{-2mm}
\end{table}

\textbf{Composition-1k.}
We first compare our MatteFormer with state-of-the-art models on Composition-1k dataset. \cref{tab:result_composition1k} tabulates the quantitative results of recent methods and shows that our method outperforms others achieving a new state-of-the-art performance.
We set\cite{yu2021mask} as our strong baseline for comparison because we follow many of its experimental details. However, in original paper, it use a binary mask as an extra input instead of trimap. For fair comparison, we retrain the same model and more larger model (with ResNet-50~\cite{he2016deep} backbone) with a trimap and set them as baselines for comparison (marked as * in \cref{tab:result_composition1k}).
Some reasons we do not take~\cite{sun2021semantic} and~\cite{liu2021tripartite} as our baseline are that~\cite{sun2021semantic} used classes of matting patterns as additional semantic information and~\cite{liu2021tripartite} showed slightly less performance than our base model, a reproduction of MG Matting with the trimap input.
\cref{fig:qualitative_results} shows visual comparison results on Composition-1k among different methods demonstrating the effectiveness of our method.

\begin{table}[t!]
  \centering
   \begin{adjustbox}{width = \linewidth}
  \begin{tabular}{l | c c c c}
    \toprule
    Method & SAD & \makecell{MSE \\ ($10^{-3}$)} & Grad & Conn\\
    \midrule
    Learning Based Matting \cite{zheng2009learning} & 105.0 & 21 & 94.2 & 110.4 \\
    Closed-Form Matting  \cite{levin2007closed} & 105.7 & 23 & 91.8 & 114.6 \\
    KNN Matting \cite{chen2013knn} & 116.7 & 25 & 103.2 & 121.5 \\
    Deep Image Matting \cite{xu2017deep} & 47.6 & 9 & 43.3 & 55.9 \\
    HAttMatting \cite{qiao2020attention} & 49.0 & 9 & 41.6 & 49.9 \\
    \hline
    MG Matting-trimap* & 23.9 & 7.4 & 14.0 & 22.4 \\
    \bf{Ours (MatteFormer)} & \bf{21.9} & \bf{6.6} & \bf{11.2} & \bf{20.5} \\
    \bottomrule
  \end{tabular}
  \end{adjustbox}
  \caption{Results on Distinctions-646 test set. * denotes the baseline of comparison which is our reproduction of MG Matting with trimap input.}
  \label{tab:result_distinction646}
  \vspace{-2mm}
\end{table}

\textbf{Distinctions-646.}
In the case of Distinctions-646, it is hard to fairly compare with previously reported results because it does not offer official trimaps for the test set.
We first produce the trimap from the ground truth alpha matte by binarizing the foreground with a threshold and randomly dilating it.
We train both the baseline (MG Matting marked as *) described above and our MatteFormer on Distinctions-646, and then evaluate them with the same testing environment. The results are shown in the last two rows of \cref{tab:result_distinction646}.
The first five methods in the table are from\cite{qiao2020attention} as references, which are trained on Distinctions-646 train set.
The results show that MatteFormer has superior performance compared to the baseline model.

\subsection{Ablation Study}


\begin{table}[t!]
  \centering
   \begin{adjustbox}{width = \linewidth}
  \begin{tabular}{l | c c c c}
    \toprule
    Method & SAD & \makecell{MSE \\ ($10^{-3}$)} & Grad & Conn\\
    \midrule
    Baseline 
    (no prior-token) & 26.43 & 5.20 & 9.57 & 21.89 \\
    \hline
    Baseline & & & & \\
    + GAP prior-token  & 25.30 & 4.72 & 9.63 & 20.61 \\
    \hline
    Baseline & & & & \\
    + uk prior-token  & 24.70 & 4.46 & 9.10 & 19.73 \\
    + uk/fg/bg prior-token & 24.19 & 4.05 & 8.72 & 19.19 \\
    + prior-memory (MatteFormer) & 23.80 & 4.03 & 8.68 & 18.90 \\
    \bottomrule
  \end{tabular}
  \end{adjustbox}
  \caption{Ablation study on the usage of prior-tokens and prior-memory. Baseline uses a Swin Transformer Tiny model as its encoder without prior-token. Results are on the Composition-1k.}
  \label{tab:ablation}
  \vspace{-2mm}
\end{table}

\textbf{{Prior-tokens and Prior-memory}. }
In MatteFormer, the building block of each encoder stage is the PAST block which is a variant version of Swin Transformer block, as described in \cref{sec:PAST}. To show how the PAST block contributes to improving performance, we conduct an ablation study. We start from a baseline model in which we set the encoder as a pure Swin Transformer Tiny model without prior-token. The decoder and shortcuts of the baseline are the same as MatteFormer.

In the PA-WSA (Prior-Attentive Window Self-Attention) layer, a token in the local window can attend not only to the in-window spatial-tokens but also to the global prior-tokens. 
As a global prior, we first use an averaged token of all spatial-tokens (Global Average Pooling (GAP)-token). In this setting, we do not use any trimap information in self-attention layer.
Next, we use unknown prior-token as a global prior. 
Further, we use all 3 prior-tokens; foreground, background and unknown prior-token.
Finally, our proposed MatteFormer model uses all prior-tokens and introduces prior-memory which make it possible to access all prior-tokens generated from the previous blocks.

The results are shown in \cref{tab:ablation}.
The baseline which uses no prior-token performs worst in all metrics as expected.
When using unknown prior-token, the metric values were lower than when using GAP-token as a global prior. This implies that the prior-token generated via the trimap delivers more useful information than the simple GAP-token.
In the case of using 3 prior-tokens (foreground, background, unknown), it shows better performance than using 1 prior-token (unknown prior-token) only. 
This suggests that a token in a local window can refer all three prior-tokens properly in the self-attention layer. Introducing prior-memory also shows a improvement of performance.
Prior-tokens from the previous blocks contribute to making better representation in the PA-WSA layer of the current block.


\begin{figure}[t]
  \centering
  \begin{subfigure}{1.0\linewidth}
    \centering
    \includegraphics[width=0.8\linewidth]{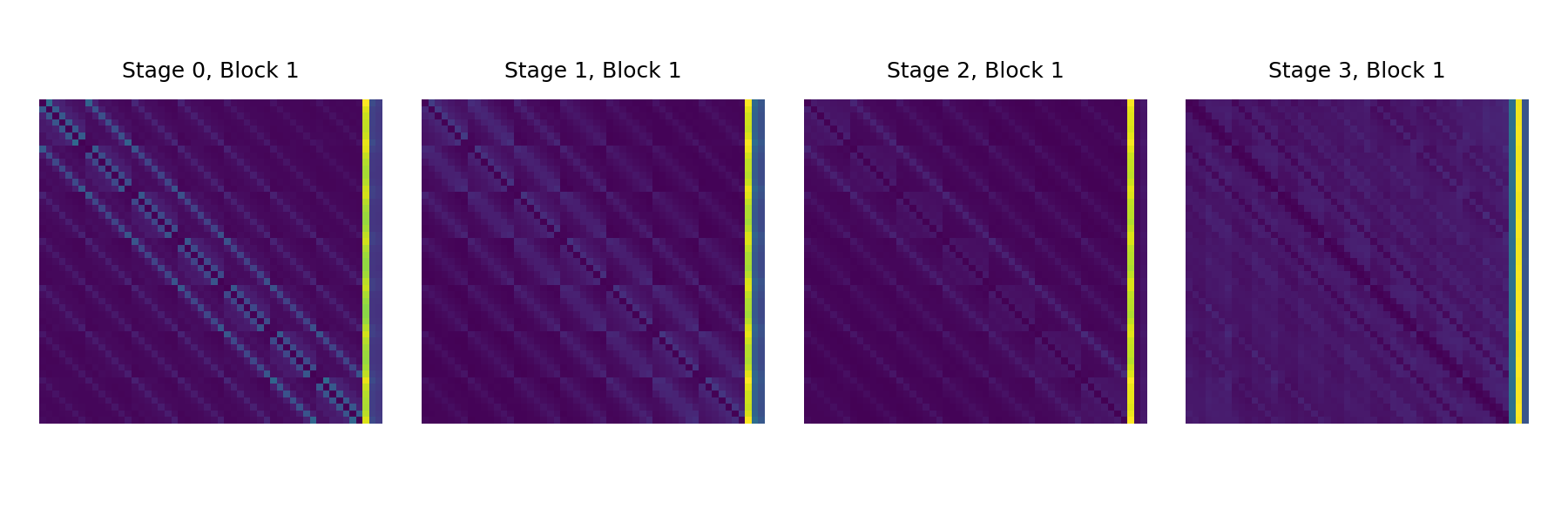}
    \vspace{-2mm}
    \caption{Mean attention maps are averaged on multi-heads. We plot only the second block (block index 1) of each stage to show in a simple.}
    \label{fig:attn_a}
  \end{subfigure}  \hfill
  
  \begin{subfigure}{1.0\linewidth}
    \centering
    \includegraphics[width=0.8\linewidth]{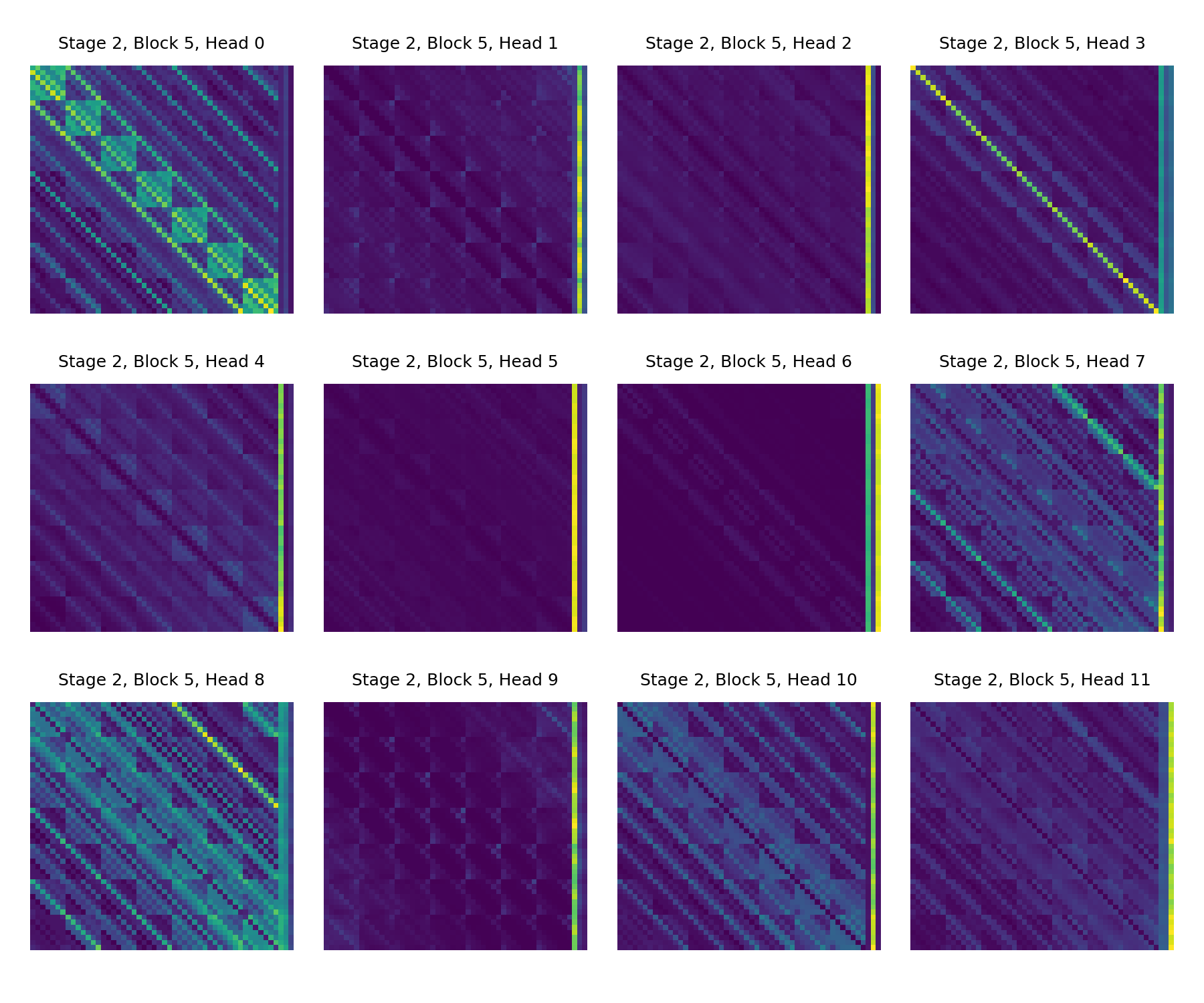}      
    \vspace{-2mm}
    \caption{Multi-head attention maps in the PA-WSA layer. We plot only the last block (block index 5) of stage index 2 to show examples in a simple. The self-attention layer has 12 multi-heads. }
    \label{fig:attn_b}
  \end{subfigure}  \hfill
  
  \begin{subfigure}{1.0\linewidth}
    \centering
    \includegraphics[width=0.7\linewidth]{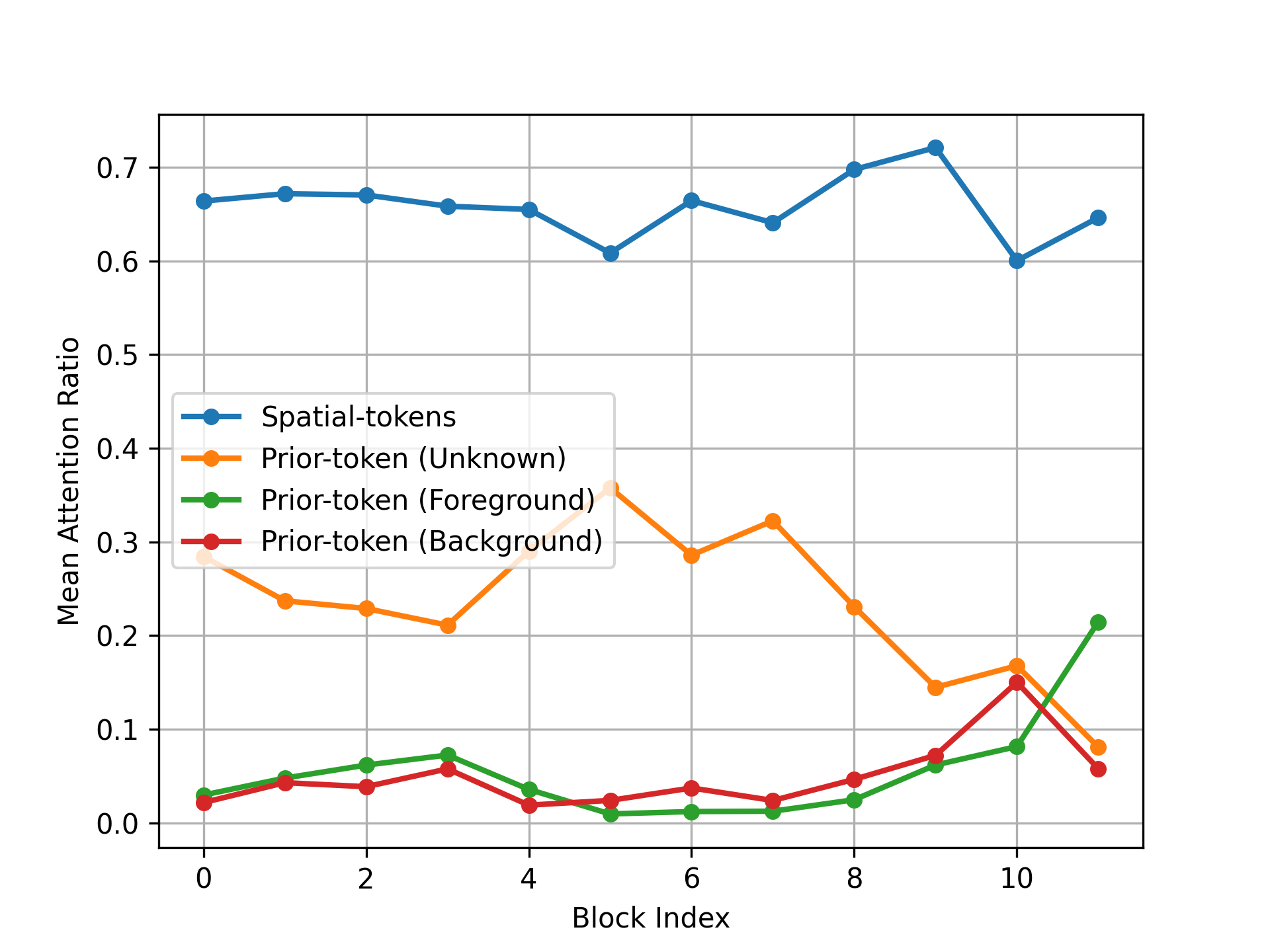}      
    \caption{Mean attention ratios on spatial-tokens and prior-tokens across all blocks.}
    \label{fig:attn_c}
  \end{subfigure}  \hfill
  
  \caption{Ablation study of attention map in PA-WSA layer. Mean attention maps and mean attention ratios show how much attention is on both spatial-tokens and prior-tokens.}
  \label{fig:attn}
  \vspace{-2mm}
\end{figure}

\vspace{2mm}
\textbf{Visualization of attention maps in PA-WSA layer. }
In this subsection, we demonstrate that the PA-WSA layer indeed has a prior-attentive property. In \cref{fig:attn}, we visualize the mean attention maps, multi-head attention maps and the mean attention ratios between local spatial-tokens and global prior-tokens. We study both of them on 50 test images with different foregrounds. We averaged the attention maps over all windows and samples. For simplicity, we conduct the ablation study on the model without prior-memory.

In \cref{fig:attn_a}, we first visualize the mean attention maps of PA-WSA layers. Since local window size is set to 7, y-axis represents 49 query spatial-tokens in a local window. Meanwhile, x-axis represents 49 spatial-tokens and 3 prior-tokens (unknown, foreground and background token in order). We can observe that the prior-token regions (last three columns) in attention maps are activated. It means that one token can attend to the prior-tokens in a self-attention layer to refer to global priors as we intended.

\cref{fig:attn_b} shows the attention maps of each head. We can see that each head has a different pattern of attention, especially in terms of prior-token usage. For example, head 4 and 5 mainly use the unknown prior-token and head 6 focuses on both unknown and background prior-tokens. Meanwhile, head 0 attends more on the spatial-tokens rather than the prior-tokens. Head 3 uses all three prior-tokens together with the spatial-tokens.

In \cref{fig:attn_c}, we visualize mean attention ratios on local spatial-tokens and global prior-tokens over all blocks, quantitatively showing that how much attention is on each prior-token. There are four lines in \cref{fig:attn_c}, one is the summation of attention ratios over local spatial-tokens, and the three others are the ratios of prior-tokens.
We observe that prior-tokens are used in the PA-WSA layers across all blocks as we have expected.
Tokens tend to attend more on the unknown prior-token than the known prior-tokens (foreground and background), which means that the unknown region is more informative to predict the alpha matte.

\label{sec:ASPP}

\begin{table}
   \begin{adjustbox}{width = \linewidth}
   \begin{tabular}{l | c c c c}
        \toprule
        \multirow{2}{*}{Methods} & \multicolumn{2}{c}{w ASPP} & \multicolumn{2}{c}{w/o ASPP} \\
        & SAD & \makecell{MSE \\ ($10^{-3}$)} & SAD & \makecell{MSE \\ ($10^{-3}$)} \\
        \hline
        Baseline (no prior-token) & 26.97 & 5.35 & 26.43 & 5.20 \\
        + u.k prior-token & 25.60 & 4.68 & 24.71 & 4.46 \\
        + u.k/f.g/b.g prior-token & 25.52 & 4.38 & 24.19 & 4.05 \\
        + prior-memory (MatteFormer) & 25.15 & 4.30 & 23.80 & 4.03 \\
        \bottomrule
    \end{tabular}
    \end{adjustbox}
    \caption{Ablation study on the usage of ASPP (Atrous Spatial Pyramid Pooling) on MatteFormer.}
    \label{tab:aspp}
\end{table}

\vspace{2mm}
\textbf{ASPP (Atrous Spatial Pyramid Pooling).} 
Many state-of-the-art image matting models~\cite{qiao2020attention, lin2021real, yu2021mask, sun2021semantic, liu2021tripartite} and many semantic segmentation methods use ASPP (Atrous Spatial Pyramid Pooling)~\cite{chen2017deeplab} between encoder and decoder to grow receptive fields for global representation. ASPP employs multiple atrous convolution filters with various atrous rates to capture spatially far-away context features. 

However, we discover that ASPP is rather hinder the performance in our model. We evaluate MatteFormer on the Composition-1k both with ASPP and without ASPP in \cref{tab:aspp}. All models with no ASPP show better performances than those with ASPP. Since our encoder is based on the transformer, MatteFormer already has a larger global receptive field than CNN-based models. As forcibly increasing receptive field with pre-defined atrous convolution, ASPP is redundant in our model. So, we do not use ASPP in our model unlike other recent methods. 


\begin{table}
  \centering
  \begin{adjustbox}{width = \linewidth}
  \begin{tabular}{l | c c c c}
    \toprule
    Method & Params & FLOPs & SAD & \makecell{MSE \\ ($10^{-3}$)}\\
    \midrule
    MG Matting-trimap* & 29.7M & 45.7G & 28.89 & 5.73 \\
    MG Matting-trimap,res50* & 52.7M & 58.9G & 28.35 & 5.42 \\
    \midrule
    Baseline (no prior-token) & 44.8M & 55.9G & 26.43 & 5.20 \\
    \hline
    Ours (MatteFormer) & 44.8M & 57.2G & 23.80 & 4.03 \\
    \bottomrule
  \end{tabular}
  \end{adjustbox}
  \caption{Parameters and FLOPs. }
  \label{tab:paramflop}
  \vspace{-2mm}
\end{table}

\vspace{2mm}
\textbf{Parameters and FLOPs.}
In \cref{tab:paramflop}, we compare the number of parameters and FLOPs of our MatteFormer to those of baseline models, for demonstrating the effectiveness of our method. Input image size is assumed to be 512 for calculating FLOPs. SAD and MSE are calculated on the Composition-1k test set for comparison.
Note that the baseline (no prior-token) uses a pure Swin Transformer Tiny model as its backbone with shortcut layers and uses no prior-token. Also, it does not use the ASPP module unlike MG Matting-trimap* and MG Matting-trimap,res50*.

First, our baseline model has fewer parameters/FLOPs than MG Matting-trimap,res50* model, but shows better performance. It means that transformer-based architecture works effectively on the image matting problem. 
Next, comparing the baseline and our MatteFormer, we can observe a quite large gap on evaluation scores with a little increase of parameters and FLOPs. Consequently, \cref{tab:paramflop} suggests that our proposed method, using prior-tokens and prior-memory, effectively works on the our transformer-based architecture.

\section{Conclusion}

In this work, we propose MatteFormer, a simple yet effective model with a modified Swin Transformer block called PAST (Prior-Attentive Swin Transformer) block for the image matting problem.
We introduce prior-tokens which represent the context of global regions separated by the given trimap. In our PAST block, prior-tokens are used as global priors with local spatial-tokens, participating at the self-attention mechanism in the PA-WSA (Prior-Attentive Window Self-Attention) layer. So, one token can attend to both local spatial-tokens and global prior-tokens.
We evaluate MatteFormer on the common datasets of the image matting problem. The experimental results show that our method achieves state-of-the-art performance.
We hope that our MatteFormer can serve as a strong baseline for future image matting models based on a transformer architecture.
One limitation of our work is that it mainly focuses on the encoder structure and the trimap-based approach. In the future work, we would like to design a fully transformer-based model with prior-tokens and also extend our model to trimap-free methods.

\clearpage
{\small
\bibliographystyle{ieee_fullname}
\bibliography{egbib}

\begin{thebibliography}{10}\itemsep=-1pt

\bibitem{beal2020toward}
Josh Beal, Eric Kim, Eric Tzeng, Dong~Huk Park, Andrew Zhai, and Dmitry
  Kislyuk.
\newblock Toward transformer-based object detection.
\newblock {\em arXiv preprint arXiv:2012.09958}, 2020.

\bibitem{cai2019disentangled}
Shaofan Cai, Xiaoshuai Zhang, Haoqiang Fan, Haibin Huang, Jiangyu Liu, Jiaming
  Liu, Jiaying Liu, Jue Wang, and Jian Sun.
\newblock Disentangled image matting.
\newblock In {\em Proceedings of the IEEE/CVF International Conference on
  Computer Vision}, pages 8819--8828, 2019.

\bibitem{carion2020end}
Nicolas Carion, Francisco Massa, Gabriel Synnaeve, Nicolas Usunier, Alexander
  Kirillov, and Sergey Zagoruyko.
\newblock End-to-end object detection with transformers.
\newblock In {\em European Conference on Computer Vision}, pages 213--229.
  Springer, 2020.

\bibitem{chen2021pre}
Hanting Chen, Yunhe Wang, Tianyu Guo, Chang Xu, Yiping Deng, Zhenhua Liu, Siwei
  Ma, Chunjing Xu, Chao Xu, and Wen Gao.
\newblock Pre-trained image processing transformer.
\newblock In {\em Proceedings of the IEEE/CVF Conference on Computer Vision and
  Pattern Recognition}, pages 12299--12310, 2021.

\bibitem{chen2017deeplab}
Liang-Chieh Chen, George Papandreou, Iasonas Kokkinos, Kevin Murphy, and Alan~L
  Yuille.
\newblock Deeplab: Semantic image segmentation with deep convolutional nets,
  atrous convolution, and fully connected crfs.
\newblock {\em IEEE transactions on pattern analysis and machine intelligence},
  40(4):834--848, 2017.

\bibitem{chen2013knn}
Qifeng Chen, Dingzeyu Li, and Chi-Keung Tang.
\newblock Knn matting.
\newblock {\em IEEE transactions on pattern analysis and machine intelligence},
  35(9):2175--2188, 2013.

\bibitem{chu2021we}
Xiangxiang Chu, Bo Zhang, Zhi Tian, Xiaolin Wei, and Huaxia Xia.
\newblock Do we really need explicit position encodings for vision
  transformers?
\newblock {\em arXiv e-prints}, pages arXiv--2102, 2021.

\bibitem{chuang2001bayesian}
Yung-Yu Chuang, Brian Curless, David~H Salesin, and Richard Szeliski.
\newblock A bayesian approach to digital matting.
\newblock In {\em Proceedings of the 2001 IEEE Computer Society Conference on
  Computer Vision and Pattern Recognition. CVPR 2001}, volume~2, pages II--II.
  IEEE, 2001.

\bibitem{deng2009imagenet}
Jia Deng, Wei Dong, Richard Socher, Li-Jia Li, Kai Li, and Li Fei-Fei.
\newblock Imagenet: A large-scale hierarchical image database.
\newblock In {\em 2009 IEEE conference on computer vision and pattern
  recognition}, pages 248--255. Ieee, 2009.

\bibitem{dong2021cswin}
Xiaoyi Dong, Jianmin Bao, Dongdong Chen, Weiming Zhang, Nenghai Yu, Lu Yuan,
  Dong Chen, and Baining Guo.
\newblock Cswin transformer: A general vision transformer backbone with
  cross-shaped windows.
\newblock {\em arXiv preprint arXiv:2107.00652}, 2021.

\bibitem{dosovitskiy2020image}
Alexey Dosovitskiy, Lucas Beyer, Alexander Kolesnikov, Dirk Weissenborn,
  Xiaohua Zhai, Thomas Unterthiner, Mostafa Dehghani, Matthias Minderer, Georg
  Heigold, Sylvain Gelly, et~al.
\newblock An image is worth 16x16 words: Transformers for image recognition at
  scale.
\newblock {\em arXiv preprint arXiv:2010.11929}, 2020.

\bibitem{everingham2010pascal}
Mark Everingham, Luc Van~Gool, Christopher~KI Williams, John Winn, and Andrew
  Zisserman.
\newblock The pascal visual object classes (voc) challenge.
\newblock {\em International journal of computer vision}, 88(2):303--338, 2010.

\bibitem{gastal2010shared}
Eduardo~SL Gastal and Manuel~M Oliveira.
\newblock Shared sampling for real-time alpha matting.
\newblock In {\em Computer Graphics Forum}, volume~29, pages 575--584. Wiley
  Online Library, 2010.

\bibitem{han2021transformer}
Kai Han, An Xiao, Enhua Wu, Jianyuan Guo, Chunjing Xu, and Yunhe Wang.
\newblock Transformer in transformer.
\newblock {\em arXiv preprint arXiv:2103.00112}, 2021.

\bibitem{he2011global}
Kaiming He, Christoph Rhemann, Carsten Rother, Xiaoou Tang, and Jian Sun.
\newblock A global sampling method for alpha matting.
\newblock In {\em CVPR 2011}, pages 2049--2056. IEEE, 2011.

\bibitem{he2010fast}
Kaiming He, Jian Sun, and Xiaoou Tang.
\newblock Fast matting using large kernel matting laplacian matrices.
\newblock In {\em 2010 IEEE Computer Society Conference on Computer Vision and
  Pattern Recognition}, pages 2165--2172. IEEE, 2010.

\bibitem{he2016deep}
Kaiming He, Xiangyu Zhang, Shaoqing Ren, and Jian Sun.
\newblock Deep residual learning for image recognition.
\newblock In {\em Proceedings of the IEEE conference on computer vision and
  pattern recognition}, pages 770--778, 2016.

\bibitem{hou2019context}
Qiqi Hou and Feng Liu.
\newblock Context-aware image matting for simultaneous foreground and alpha
  estimation.
\newblock In {\em Proceedings of the IEEE/CVF International Conference on
  Computer Vision}, pages 4130--4139, 2019.

\bibitem{lee2011nonlocal}
Philip Lee and Ying Wu.
\newblock Nonlocal matting.
\newblock In {\em CVPR 2011}, pages 2193--2200. IEEE, 2011.

\bibitem{levin2007closed}
Anat Levin, Dani Lischinski, and Yair Weiss.
\newblock A closed-form solution to natural image matting.
\newblock {\em IEEE transactions on pattern analysis and machine intelligence},
  30(2):228--242, 2007.

\bibitem{levin2008spectral}
Anat Levin, Alex Rav-Acha, and Dani Lischinski.
\newblock Spectral matting.
\newblock {\em IEEE transactions on pattern analysis and machine intelligence},
  30(10):1699--1712, 2008.

\bibitem{li2020natural}
Yaoyi Li and Hongtao Lu.
\newblock Natural image matting via guided contextual attention.
\newblock In {\em Proceedings of the AAAI Conference on Artificial
  Intelligence}, volume~34, pages 11450--11457, 2020.

\bibitem{li2021localvit}
Yawei Li, Kai Zhang, Jiezhang Cao, Radu Timofte, and Luc Van~Gool.
\newblock Localvit: Bringing locality to vision transformers.
\newblock {\em arXiv preprint arXiv:2104.05707}, 2021.

\bibitem{liang2021swinir}
Jingyun Liang, Jiezhang Cao, Guolei Sun, Kai Zhang, Luc Van~Gool, and Radu
  Timofte.
\newblock Swinir: Image restoration using swin transformer.
\newblock In {\em Proceedings of the IEEE/CVF International Conference on
  Computer Vision}, pages 1833--1844, 2021.

\bibitem{lin2021real}
Shanchuan Lin, Andrey Ryabtsev, Soumyadip Sengupta, Brian~L Curless, Steven~M
  Seitz, and Ira Kemelmacher-Shlizerman.
\newblock Real-time high-resolution background matting.
\newblock In {\em Proceedings of the IEEE/CVF Conference on Computer Vision and
  Pattern Recognition}, pages 8762--8771, 2021.

\bibitem{lin2014microsoft}
Tsung-Yi Lin, Michael Maire, Serge Belongie, James Hays, Pietro Perona, Deva
  Ramanan, Piotr Doll{\'a}r, and C~Lawrence Zitnick.
\newblock Microsoft coco: Common objects in context.
\newblock In {\em European conference on computer vision}, pages 740--755.
  Springer, 2014.

\bibitem{liu2021towards}
Chang Liu, Henghui Ding, and Xudong Jiang.
\newblock Towards enhancing fine-grained details for image matting.
\newblock In {\em Proceedings of the IEEE/CVF Winter Conference on Applications
  of Computer Vision}, pages 385--393, 2021.

\bibitem{liu2021transformer}
Yun Liu, Guolei Sun, Yu Qiu, Le Zhang, Ajad Chhatkuli, and Luc Van~Gool.
\newblock Transformer in convolutional neural networks.
\newblock {\em arXiv preprint arXiv:2106.03180}, 2021.

\bibitem{liu2021tripartite}
Yuhao Liu, Jiake Xie, Xiao Shi, Yu Qiao, Yujie Huang, Yong Tang, and Xin Yang.
\newblock Tripartite information mining and integration for image matting.
\newblock In {\em Proceedings of the IEEE/CVF International Conference on
  Computer Vision}, pages 7555--7564, 2021.

\bibitem{liu2021swin}
Ze Liu, Yutong Lin, Yue Cao, Han Hu, Yixuan Wei, Zheng Zhang, Stephen Lin, and
  Baining Guo.
\newblock Swin transformer: Hierarchical vision transformer using shifted
  windows.
\newblock {\em arXiv preprint arXiv:2103.14030}, 2021.

\bibitem{lu2019indices}
Hao Lu, Yutong Dai, Chunhua Shen, and Songcen Xu.
\newblock Indices matter: Learning to index for deep image matting.
\newblock In {\em Proceedings of the IEEE/CVF International Conference on
  Computer Vision}, pages 3266--3275, 2019.

\bibitem{lu2021efficient}
Zhisheng Lu, Hong Liu, Juncheng Li, and Linlin Zhang.
\newblock Efficient transformer for single image super-resolution.
\newblock {\em arXiv preprint arXiv:2108.11084}, 2021.

\bibitem{lutz2018alphagan}
Sebastian Lutz, Konstantinos Amplianitis, and Aljosa Smolic.
\newblock Alphagan: Generative adversarial networks for natural image matting.
\newblock {\em arXiv preprint arXiv:1807.10088}, 2018.

\bibitem{qiao2020attention}
Yu Qiao, Yuhao Liu, Xin Yang, Dongsheng Zhou, Mingliang Xu, Qiang Zhang, and
  Xiaopeng Wei.
\newblock Attention-guided hierarchical structure aggregation for image
  matting.
\newblock In {\em Proceedings of the IEEE/CVF Conference on Computer Vision and
  Pattern Recognition}, pages 13676--13685, 2020.

\bibitem{ryoo2021tokenlearner}
Michael~S Ryoo, AJ Piergiovanni, Anurag Arnab, Mostafa Dehghani, and Anelia
  Angelova.
\newblock Tokenlearner: What can 8 learned tokens do for images and videos?
\newblock {\em arXiv preprint arXiv:2106.11297}, 2021.

\bibitem{sengupta2020background}
Soumyadip Sengupta, Vivek Jayaram, Brian Curless, Steven~M Seitz, and Ira
  Kemelmacher-Shlizerman.
\newblock Background matting: The world is your green screen.
\newblock In {\em Proceedings of the IEEE/CVF Conference on Computer Vision and
  Pattern Recognition}, pages 2291--2300, 2020.

\bibitem{shahrian2013improving}
Ehsan Shahrian, Deepu Rajan, Brian Price, and Scott Cohen.
\newblock Improving image matting using comprehensive sampling sets.
\newblock In {\em Proceedings of the IEEE Conference on Computer Vision and
  Pattern Recognition}, pages 636--643, 2013.

\bibitem{sun2004poisson}
Jian Sun, Jiaya Jia, Chi-Keung Tang, and Heung-Yeung Shum.
\newblock Poisson matting.
\newblock In {\em ACM SIGGRAPH 2004 Papers}, pages 315--321. 2004.

\bibitem{sun2021semantic}
Yanan Sun, Chi-Keung Tang, and Yu-Wing Tai.
\newblock Semantic image matting.
\newblock In {\em Proceedings of the IEEE/CVF Conference on Computer Vision and
  Pattern Recognition}, pages 11120--11129, 2021.

\bibitem{tang2019learning}
Jingwei Tang, Yagiz Aksoy, Cengiz Oztireli, Markus Gross, and Tunc~Ozan Aydin.
\newblock Learning-based sampling for natural image matting.
\newblock In {\em Proceedings of the IEEE/CVF Conference on Computer Vision and
  Pattern Recognition}, pages 3055--3063, 2019.

\bibitem{touvron2021training}
Hugo Touvron, Matthieu Cord, Matthijs Douze, Francisco Massa, Alexandre
  Sablayrolles, and Herv{\'e} J{\'e}gou.
\newblock Training data-efficient image transformers \& distillation through
  attention.
\newblock In {\em International Conference on Machine Learning}, pages
  10347--10357. PMLR, 2021.

\bibitem{vaswani2021scaling}
Ashish Vaswani, Prajit Ramachandran, Aravind Srinivas, Niki Parmar, Blake
  Hechtman, and Jonathon Shlens.
\newblock Scaling local self-attention for parameter efficient visual
  backbones.
\newblock In {\em Proceedings of the IEEE/CVF Conference on Computer Vision and
  Pattern Recognition}, pages 12894--12904, 2021.

\bibitem{vaswani2017attention}
Ashish Vaswani, Noam Shazeer, Niki Parmar, Jakob Uszkoreit, Llion Jones,
  Aidan~N Gomez, {\L}ukasz Kaiser, and Illia Polosukhin.
\newblock Attention is all you need.
\newblock In {\em Advances in neural information processing systems}, pages
  5998--6008, 2017.

\bibitem{wan2021high}
Ziyu Wan, Jingbo Zhang, Dongdong Chen, and Jing Liao.
\newblock High-fidelity pluralistic image completion with transformers.
\newblock {\em arXiv preprint arXiv:2103.14031}, 2021.

\bibitem{wang2007optimized}
Jue Wang and Michael~F Cohen.
\newblock Optimized color sampling for robust matting.
\newblock In {\em 2007 IEEE Conference on Computer Vision and Pattern
  Recognition}, pages 1--8. IEEE, 2007.

\bibitem{wang2021pyramid}
Wenhai Wang, Enze Xie, Xiang Li, Deng-Ping Fan, Kaitao Song, Ding Liang, Tong
  Lu, Ping Luo, and Ling Shao.
\newblock Pyramid vision transformer: A versatile backbone for dense prediction
  without convolutions.
\newblock {\em arXiv preprint arXiv:2102.12122}, 2021.

\bibitem{wang2021anchor}
Yingming Wang, Xiangyu Zhang, Tong Yang, and Jian Sun.
\newblock Anchor detr: Query design for transformer-based detector.
\newblock {\em arXiv preprint arXiv:2109.07107}, 2021.

\bibitem{wu2020visual}
Bichen Wu, Chenfeng Xu, Xiaoliang Dai, Alvin Wan, Peizhao Zhang, Zhicheng Yan,
  Masayoshi Tomizuka, Joseph Gonzalez, Kurt Keutzer, and Peter Vajda.
\newblock Visual transformers: Token-based image representation and processing
  for computer vision.
\newblock {\em arXiv preprint arXiv:2006.03677}, 2020.

\bibitem{wu2021cvt}
Haiping Wu, Bin Xiao, Noel Codella, Mengchen Liu, Xiyang Dai, Lu Yuan, and Lei
  Zhang.
\newblock Cvt: Introducing convolutions to vision transformers.
\newblock {\em arXiv preprint arXiv:2103.15808}, 2021.

\bibitem{xie2021segformer}
Enze Xie, Wenhai Wang, Zhiding Yu, Anima Anandkumar, Jose~M Alvarez, and Ping
  Luo.
\newblock Segformer: Simple and efficient design for semantic segmentation with
  transformers.
\newblock {\em arXiv preprint arXiv:2105.15203}, 2021.

\bibitem{xu2017deep}
Ning Xu, Brian Price, Scott Cohen, and Thomas Huang.
\newblock Deep image matting.
\newblock In {\em Proceedings of the IEEE Conference on Computer Vision and
  Pattern Recognition}, pages 2970--2979, 2017.

\bibitem{yang2021focal}
Jianwei Yang, Chunyuan Li, Pengchuan Zhang, Xiyang Dai, Bin Xiao, Lu Yuan, and
  Jianfeng Gao.
\newblock Focal self-attention for local-global interactions in vision
  transformers.
\newblock {\em arXiv preprint arXiv:2107.00641}, 2021.

\bibitem{yu2020high}
Haichao Yu, Ning Xu, Zilong Huang, Yuqian Zhou, and Humphrey Shi.
\newblock High-resolution deep image matting.
\newblock {\em arXiv preprint arXiv:2009.06613}, 2020.

\bibitem{yu2021mask}
Qihang Yu, Jianming Zhang, He Zhang, Yilin Wang, Zhe Lin, Ning Xu, Yutong Bai,
  and Alan Yuille.
\newblock Mask guided matting via progressive refinement network.
\newblock In {\em Proceedings of the IEEE/CVF Conference on Computer Vision and
  Pattern Recognition}, pages 1154--1163, 2021.

\bibitem{yuan2021tokens}
Li Yuan, Yunpeng Chen, Tao Wang, Weihao Yu, Yujun Shi, Zihang Jiang, Francis~EH
  Tay, Jiashi Feng, and Shuicheng Yan.
\newblock Tokens-to-token vit: Training vision transformers from scratch on
  imagenet.
\newblock {\em arXiv preprint arXiv:2101.11986}, 2021.

\bibitem{zhang2021multi}
Pengchuan Zhang, Xiyang Dai, Jianwei Yang, Bin Xiao, Lu Yuan, Lei Zhang, and
  Jianfeng Gao.
\newblock Multi-scale vision longformer: A new vision transformer for
  high-resolution image encoding.
\newblock {\em arXiv preprint arXiv:2103.15358}, 2021.

\bibitem{zhang2019late}
Yunke Zhang, Lixue Gong, Lubin Fan, Peiran Ren, Qixing Huang, Hujun Bao, and
  Weiwei Xu.
\newblock A late fusion cnn for digital matting.
\newblock In {\em Proceedings of the IEEE/CVF Conference on Computer Vision and
  Pattern Recognition}, pages 7469--7478, 2019.

\bibitem{zheng2021rethinking}
Sixiao Zheng, Jiachen Lu, Hengshuang Zhao, Xiatian Zhu, Zekun Luo, Yabiao Wang,
  Yanwei Fu, Jianfeng Feng, Tao Xiang, Philip~HS Torr, et~al.
\newblock Rethinking semantic segmentation from a sequence-to-sequence
  perspective with transformers.
\newblock In {\em Proceedings of the IEEE/CVF Conference on Computer Vision and
  Pattern Recognition}, pages 6881--6890, 2021.

\bibitem{zheng2009learning}
Yuanjie Zheng and Chandra Kambhamettu.
\newblock Learning based digital matting.
\newblock In {\em 2009 IEEE 12th international conference on computer vision},
  pages 889--896. IEEE, 2009.

\end{thebibliography}
}

\end{document}